\documentclass[a4paper]{article}

\usepackage{authblk}
\usepackage{algorithm}
\usepackage{algorithmic}
\usepackage{multirow}
\usepackage{graphicx}
\usepackage{booktabs}
\usepackage{epsfig}
\usepackage{epstopdf}
\usepackage{amsmath}
\usepackage{caption}
\usepackage{subcaption}
\usepackage{amsfonts}
\usepackage{wasysym}
\usepackage{paralist}
\usepackage{url}
\usepackage{float}
\usepackage{booktabs}
\usepackage{todonotes}
\usepackage{microtype}
\usepackage{tikz}
\usepackage{tikz-qtree}
\usepackage{fancyhdr}
\usepackage{textcomp}

\fancypagestyle{firstpage}{%
	\fancyhf{}
	\chead{{\scriptsize This is a pre-print version of an article accepted for publication at the Genetic and Evolutionary Computation Conference -- GECCO 2020}}}

\fancypagestyle{otherpages}{%
	\fancyhf{}
	\chead{{\scriptsize This is a pre-print version of an article accepted for publication at the Genetic and Evolutionary Computation Conference -- GECCO 2020}}
	\cfoot{\thepage}}

\newcommand{\R}{\mathbb{R}}
\providecommand{\keywords}[1]{\textbf{\textit{Keywords }} #1}

\begin{document}

\title{Towards an evolutionary-based approach for natural language processing}

\author[1]{Luca Manzoni}
\author[2]{Domagoj Jakobovic}
\author[3]{Luca Mariot}
\author[3]{Stjepan Picek}
\author[4]{Mauro Castelli}

\affil[1]{{\normalsize Dipartimento di Matematica e Geoscienze, Università degli Studi di Trieste, Via Valerio 12/1, 34127 Trieste, Italy} \\
	
	{\small \texttt{lmanzoni@units.it}}}

\affil[2]{{\normalsize Faculty of Electrical Engineering and Computing, University of Zagreb, Unska 3, Zagreb, Croatia} \\
	
	{\small \texttt{domagoj.jakobovic@fer.hr}}}

\affil[3]{{\normalsize Cyber Security Research Group, Delft University of Technology,, Mekelweg 2, Delft, The Netherlands} \\
	
	{\small \texttt{\{l.mariot,s.picek\}@tudelft.nl}}}

\affil[4]{{\normalsize NOVA Information Management School (NOVA IMS), Universidade Nova de Lisboa, Campus de Campolide, 1070-312, Lisbon, Portugal} \\
	
	{\small \texttt{mcastelli@novaims.unl.pt}}}
\maketitle

\begin{abstract}
    Tasks related to Natural Language Processing (NLP) have recently been the focus of a large research endeavor by the machine learning community. The increased interest in this area is mainly due to the success of deep learning methods. Genetic Programming (GP), however, was not under the spotlight with respect to NLP tasks. Here, we propose a first proof-of-concept that combines GP with the well established NLP tool word2vec for the next word prediction task. The main idea is that, once words have been moved into a vector space, traditional GP operators can successfully work on vectors, thus producing meaningful words as the output. To assess the suitability of this approach, we perform an experimental evaluation on a set of existing newspaper headlines. Individuals resulting from this (pre-)training phase can be employed as the initial population in other NLP tasks, like sentence generation, which will be the focus of future investigations, possibly employing adversarial co-evolutionary approaches.
\end{abstract}

\keywords{Genetic Programming $\cdot$ Natural Language Processing $\cdot$ Next word prediction}
\thispagestyle{firstpage}

\section{Introduction}
\label{sec:introduction}

Natural language processing (NLP) is a branch of artificial intelligence that analyzes naturally occurring texts to achieve human-like language processing for different applications~\cite{dale2000handbook}. 
With the increasing popularity of deep learning, NLP is nowadays a hot research topic in the scientific community, and several contributions (mainly based on neural models) appeared in recent years~\cite{young2018recent}.
Next word prediction is one of the most important tasks in this research field, and, given a sequence of words, it aims at predicting what word comes next. The importance of this task is motivated by the vast amount of applications in which it appears, including the composition of text and/or emails as well as ad-hoc applications designed to help people with physical disabilities to communicate~\cite{al2008application}. 
The next word prediction problem can be addressed by considering different techniques~\cite{garay2006text}. The simplest approach is the one that takes into account words and their respective frequencies~\cite{garay1994using}. In this case, when a user entered the first characters of a word, the system suggests the most probable words beginning with the same character (or characters). While this method is simple to be implemented, it relies on standard word frequencies, thus ignoring the lexicon of different users. Thus, to obtain better suggestions, it is necessary to update the system once the user has entered a significant amount of text~\cite{venkatagiri1993efficiency}. 

A different method to address the next word prediction problem takes into account the probability of appearance of each word after the one previously entered. In other words, instead of considering the simple information about words' frequencies, this approach uses a two-dimensional table containing the conditional probability of the word $w_j$ appearing after the word $w_i$ was entered. While this approach results in better prediction with respect to the one based only on words' frequencies, it has an important limitation associated with the size of the table that must be stored. Additionally, as pointed out in~\cite{garay2006text}, it is difficult to adapt the system to the user's vocabulary when the dimensions
of the table are fixed. Hence, this approach is commonly used in combination with the previous one, and the table only stores the most probable word-pairs.

A different method considers the syntactic information inherent to natural languages to address the next word prediction task~\cite{swiffin1987use}. To implement this idea, we need two pieces of information: the first one is the probability of appearance of each word, while the second is the relative probability of appearance of every syntactic category after each syntactic category. By exploiting the syntactic information associated with the language, this approach results in better predictions with respect to the ones achieved by the aforementioned methods. 
The main limitation of this approach is the presence of words with ambiguous categories, that may significantly affect the quality of the prediction as well as the subsequent predictions~\cite{garay2006text}.
A different approach is based on the analysis of sentences by the use of grammars and by applying NLP techniques to obtain the categories with the highest probability of appearance~\cite{garay1997intelligent}. 

\pagestyle{otherpages}

Nowadays, the most common and successful approaches to deal with NLP tasks are based on the use of deep learning techniques. This is mostly due to the ability of deep learning in modeling the recursivity of human language that is composed of words and sentences with a certain structure. Deep learning (especially recurrent neural models), can capture the sequence information more effectively compared to other existing techniques.
The reader is referred to~\cite{young2018recent} and~\cite{alshemali2019improving} for a complete review of the existing deep learning-based approaches in the field of NLP.

Despite the existence of different techniques dedicated to the next word prediction task, the use of evolutionary algorithms was not fully explored. In particular, Genetic Programming (GP) was applied to different tasks in the context of NLP~\cite{araujo2019genetic}, but no specific effort was dedicated to this prediction task.

Considering the ability of GP to address problems over different domains and its ability to explore search spaces characterized by a high-dimensionality, in this paper, we answer this call by presenting a GP-based system for the next word prediction problem. The choice of using GP is also motivated by the fact that it can provide information that cannot be extracted from a black-box model, thus providing a more interpretable model. The ability of GP to provide human-understandable solutions is particularly useful in the context of this application because it may provide interesting hints about the nature of the language and the learning process of GP. Additionally, this task can be addressed by building rules from the set of words in input, and GP was successfully used to generate rules-based models~\cite{banzhaf2019genetic} for problems over different domains. Thus it is reasonable to investigate the potential of GP in addressing the next word prediction task.

This paper is organized as follows: related works are presented in Section~\ref{sec:related}, while the proposed method for next word prediction with GP is described in Section~\ref{sec:nextword}. The experimental settings and results are presented in Section~\ref{sec:exp} and discussed in Section~\ref{sec:disc}. Section~\ref{sec:conc} concludes the paper and highlight some possible directions for future research.

\section{Related Works}
\label{sec:related}

This section reviews the use of Genetic Programming for addressing different tasks in the field of natural language processing.
While this research field is nowadays dominated by the use of deep learning models, we decided to restrict our attention to the GP-based techniques that represent the method explored in this paper.

The first applications of GP in the field of NLP were mostly
related to the identification of the semantic structure of the language. The main reason relies on the fact that the syntax-tree representation used in GP is similar to the one commonly employed to describe the syntax of a language~\cite{araujo2019genetic}. Thus, it comes naturally to use GP for the task of identifying the syntactic structures of the language.
One of the first studies in this area was proposed in~\cite{rose1999genetic}, where the
author employed GP to address a parser failure problem in speech-to-speech machine translation.
In the same research direction, Araujo~\cite{araujo2004genetic} proposed a GP-based system for natural language parsing. The system was subsequently improved by considering a multi-objective formulation for simultaneous performance of language tagging and parsing~\cite{1010071184429744}.

Other works where GP was used for NLP tasks took into consideration the generation of regular expressions. Regular expressions are important in the field of NLP due to their ability to represent string patterns precisely.
In~\cite{bartoli2014automatic}, Bartoli and coauthors applied GP for extracting regular expressions for entity extraction applications. They designed a multi-objective GP-based system to automatically create a regular expression for a task that was represented through a set of examples. Experimental results demonstrated the suitability of the system they proposed, with the GP-based approach outperforming the existing techniques already tested on the same datasets. 
Subsequently, in~\cite{bartoli2017active}, Bartoli and coauthors described an active learning approach where the user acts as an oracle. They assessed the performance of this approach on the same datasets used in~\cite{bartoli2014automatic}, and they demonstrated that the active learning approach might reduce the computational cost by requiring a lower amount of user annotations.  

Another task where GP was employed is the search and extraction of semantic relationships in texts, which is especially relevant in the medical domain. In this domain, GP was used to identify sentences that contain descriptions of interactions between genes and proteins~\cite{bartoli2015evolutionary}. More in detail, GP was employed to obtain a model of syntax patterns composed of part-of-speech tags. The model consisted of a set of automatically learned regular expressions, following the idea developed in~\cite{bartoli2014automatic}.

Other works focused on the entity linking task, the identification of the different ways in which the same entity is mentioned in the texts. To address this task, Carvalho and coauthors~\cite{de2010genetic} used GP to find effective functions that can identify, in a data repository, entries referring to the same entity despite typos.
Experimental results demonstrated that GP was able to outperform the state-of-the-art SVM-based approach over real-world data. 
Subsequently, a combination of GP and active learning was proposed in~\cite{isele2013active} to address the same problem, and the results demonstrated the beneficial effect of combining active learning with GP.

Moving to the task of natural language generation, Manurung and coauthors~\cite{manurung2008implementation} developed a GP system for generating poetry
with a certain meter or patterns in the rhythm, while Kim and coauthors~\cite{kim2004user} developed a GP system for generating answers that an agent must provide to users' queries. These are the only GP-based contributions in the field of natural language generation, and they cover very specific domains. Additionally, in these works, GP is used to represent the grammars that must be evolved.

For a complete review of the applications of GP in the NLP field, the reader is referred to~\cite{araujo2019genetic}.

\section{Next word Prediction with GP}
\label{sec:nextword}
In this section, we describe how to adapt GP to work with words both at the input and the output levels and how to manipulate them. In particular, we tackle the problem of \emph{next word prediction}, where, given an initial sequence of words, possibly of fixed length $k$, we want to predict the next word in the sequence. Since we need to produce a word, we have to perform text generation via GP. To accomplish that, we need to tackle multiple obstacles; specifically, there are the following points to be taken care of:
\begin{itemize}
	\item \textbf{Input representation}. How can the input words be represented in a suitable way for GP?
	\item \textbf{Functional operators}. What operations can be performed on the representation of the words?
	\item \textbf{Output interpretation}. How can we decode the output of a GP individual and interpret it as a word?
\end{itemize}
These three questions must be addressed, and are all related to the same problem of deciding in which genotypic space GP has to work. In particular, the input representation will influence which kind of operations one can perform, and the operations will also dictate how the output will be decoded.

Behind our approach to all three problems lies the adoption of an \emph{embedding}, where words are embedded into a vector space, usually $\R^d$ for some dimension $d$. Among the existing embedding methods, one that has risen to prominence since its inception in 2013 is \emph{word2vec}~\cite{mikolov2013}. The algorithm takes as input a large corpus of text and produces an embedding where each word of the corpus is represented as a real-valued vector, with the property that words sharing similar contexts will be near in their vector representation. Usually, the size of the space $\R^d$ in which the words are embedded is between $100$ and $1\,000$, but we will see that, in the case of GP, a lower dimensionality appears to help. Notice that we assume that all necessary words, both for the input and the output, are present in the original corpus used to train the word2vec model. That is, the GP trees only receive inputs and produce the output only from a fixed vocabulary.

\begin{figure*}
	\centering
	\includegraphics[width=0.85\textwidth]{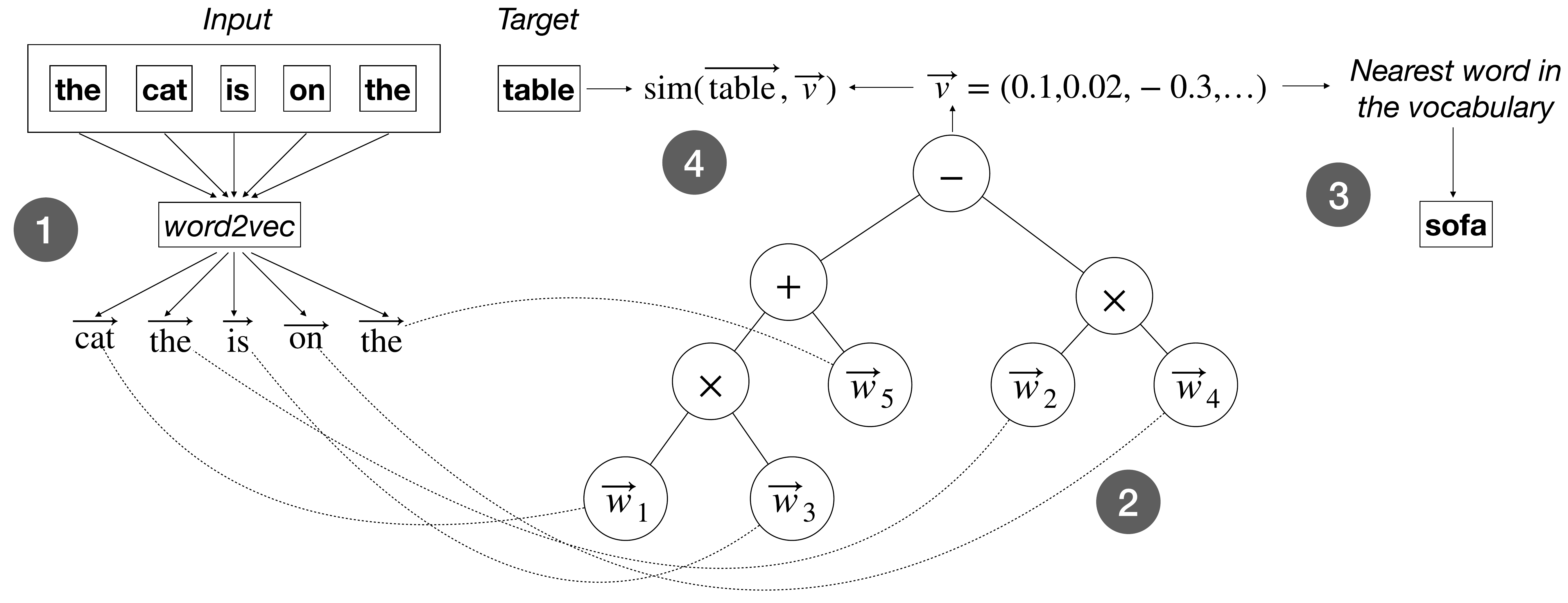}
	\caption{A graphical depiction of the process employed: conversion of the input word into vectors (1), evaluation of the GP tree (2), and conversion of the output vector into a word by finding the most similar word in the vocabulary (3). The fitness can be computed by computing the similarity between the target and output of the GP tree (4) directly.}
	\label{fig:general-method}
\end{figure*}

We are now going to detail our approach to solve all three previously highlighted obstacles. A graphical representation of the entire process is given in Figure~\ref{fig:general-method}.

\subsubsection*{Input representation}
Given a sentence of fixed length $k$ consisting of the words $w_1, \ldots, w_k$, we map each word to its corresponding vector via a word2vec embedding to interpret the sentence as an input to a GP individual. Therefore, we now have $k$ vectors $\vec{w}_1, \ldots, \vec{w}_k \in \R^d$, where $d$ is the dimension of the embedding. Without loss of generality, we consider all vectors to be of unitary length, that is $||\vec{w}_i||_2 = 1$ for all $1 \le i \le k$.

\subsubsection*{Functional operators}
Once the input encoding as unitary vectors in $\R^d$ has been defined, it is necessary to define the set of functional symbols to employ. A first requirement is that the output of each operation should itself be a unitary vector in $\R^d$. If we consider some of the common binary GP operators as used in symbolic regression, i.e., $+$, $-$, $\times$, and protected division ($\div$), they can all be extended to work on vectors. Given two vectors $\vec{u}$ and $\vec{v}$ in $\R^d$, a binary operator $\circ \colon \R \times \R \to \R$ can be extended to a binary operation $\vec{\circ} \colon \R^d \times \R^d \to \R^d$ by applying it component-wise as follows:
\begin{align*}
& \left(\vec{u} \;\vec{\circ}\; \vec{v}\right)_i = \vec{u}_i \circ \vec{v}_i & \forall 1 \le i \le n \enspace .
\end{align*}
We can further restrict the output vector to be of unitary length by normalizing it, as long as its length is not null. This allows us to define the operator $\vec{\circ}_1 \colon \R^d \times \R^d \to \R^d$ as:
\begin{align*}
& \left( \vec{u} \; \vec{\circ}_1 \; \vec{v} \right) = \frac{\left(\vec{u} \;\vec{\circ}\; \vec{v}\right)_i}{||\vec{u} \;\vec{\circ}\; \vec{v}||_2} & \forall 1 \le i \le n \enspace .
\end{align*}
The same procedure can then be extended to unary operators, such as squaring, or even to operators of higher arity, by component-wise evaluation followed by normalization.

By performing this extension process on the classical GP operators, we are then able to produce unitary vectors in $\R^d$ as output whenever vectors in $\R^d$ are given as input.

\subsubsection*{Output interpretation}
To obtain a word as the final output of our tree, one has to return from the embedding produced by word2vec to the actual set of words. Let us consider the vocabulary $V$ consisting of the $m$ distinct words $w_1, \ldots, w_m$ occurring in the corpus used to generate the embedding, and their respective vectors $\vec{V}=\{ \vec{w}_1, \ldots,\vec{w}_m\}$. The output of a GP tree is a vector $\vec{v} \in \R^d$, which might not be an element of $\vec{V}$, and, in practice, it will almost never be. Therefore, we assign to $\vec{v}$ the word corresponding to the vector $\vec{u} \in \{\vec{w}_1\ldots \vec{w}_m\}$ which is most similar to $\vec{v}$. In the case of word embeddings, a common measure of similarity is the \emph{cosine similarity}, corresponding to the cosine of the angle formed between the two  vectors $\vec{u}$ and $\vec{v}$, which is defined as:
\[
\mathrm{sim}(\vec{u}, \vec{v}) = \frac{\sum_{i=1}^{n} \vec{u}_i\vec{v}_i}{||\vec{u}||_2 ||\vec{v}||_2} \enspace .
\]
Since cosine is between $-1$ and $1$ and maximal when the two vectors coincide, we will need to select, among the words in the vocabulary, the one whose embedding has the highest similarity to $\vec{v}$.

\subsubsection*{Computation of the fitness}
Given a set of $h$ fitness cases, each one of the form $((w_1, \ldots, w_k), w_{k+1})$, where $w_1, \ldots, w_k \in V$ are the input words and $w_{k+1} \in V$ is the target output, the fitness of a GP individual $T$ is computed as the average of the cosine similarity between the output of $T$ and $\vec{w}_{k+1}$. Since we are considering cosine similarity, the fitness is to be maximized.

There is, however, a possible ambiguity in this definition: one can consider the output of $T$ either as the vector $\vec{v}$ obtained \emph{before} re-interpreting the output as a word, or the word $w_i$ from the vocabulary obtained \emph{after} ``decoding'' the vector $\vec{v}$. The two approaches might produce different results, but the second one, while arguably more ``precise'' (i.e., by considering the output of the entire system and not only of $T$), is also more computationally intensive, since a na\"ive implementation requires the computation of the cosine similarity with all words in the vocabulary for \emph{each} fitness case. Therefore, we have selected the first method to compute the fitness. That is, for each fitness case we compute the output $\vec{v}$ of $T$ and $\mathrm{sim}(\vec{v}, \vec{w}_{k+1})$. The sum of these values across all fitness cases is then averaged to produce the final fitness value of $T$.

\section{Experiments}
\label{sec:exp}
In this section, we describe the experimental setting adopted to test our GP algorithm, along with a brief description of the text dataset used to train the GP trees for the next word prediction task. We then report and discuss the results obtained during the training and the test phase.


\subsection{Experimental Setting}
\label{subsec:exp-set}
Recall that the optimization goal is to evolve a population of GP trees that take as input the first words of a sentence and predict the next word to complete the sentence. For our experiments, we considered the \emph{Million News Headlines} (MNH) dataset~\cite{rohit18}, which contains headlines collected from the Australian Broadcasting Corporation over 17 years from 2003 to 2019. In particular, for our experiments, we fixed the problem instance to headlines of six words, which amounts to 267\,292 training examples in the MNH dataset. Therefore, the prediction task of the trees evolved through GP was to generate the vector for the sixth word of these headlines by considering as input the first five words of the headline.

For the pre-training phase, we used word2vec to generate six different embeddings $E_d \subseteq \mathbb{R}^d$ of the whole MNH dataset, with dimension $d$ ranging in the set $\{10,15,20,25,50,100\}$. This was done to investigate whether the dimension of the embedding space influenced the training performance of GP. For the embedding, we employed the default parameters of word2vec, except for the size (which was a value in $\{0,15,20,25,50,100\}$), and the \texttt{min\_count} which was set to $1$, in order to insert all words in the vocabulary. Notice that this means that very infrequent words will not have a semantically ``good'' representation as vectors. Finally, the number of iterations was increased to $20$, and the number of threads to $16$.

The input variables at the terminal nodes of the trees evolved by GP represented the vectors corresponding to the first words of a sentence under the considered embedding. The set of functionals for the internal nodes of the trees included sum, subtraction, multiplication, and protected division as binary operators, and squaring and square root as unary operators. Recall that all operators are pointwise evaluated, i.e., they are applied separately on each coordinate of the input vectors.
We adopted a steady-state breeding strategy using a tournament selection operator of size $t=3$, where the two candidates with the highest fitness are mated through crossover, and the resulting child tree replaces the worst individual in the tournament after being mutated. For the variation operators, we employed simple subtree crossover, uniform crossover, size fair, one-point, and context preserving crossover, randomly selected at each generation, while for mutation, we used simple subtree mutation~\cite{poli08}. To prevent bloat, we forced a maximum depth on our GP trees equal to $5$, i.e., the number of input words. Further, we considered a population of $P=500$ individuals and a mutation probability of $p_\mu=0.3$, and we used a maximum budget of $fit=100\,000$ fitness evaluations as a termination criterion. All these parameters were selected after a preliminary tuning phase, where we observed that perturbing the above values did not yield any significant difference in performances. In particular, we detected no substantial increase in the fitness value of the best solution after the 100\, 000 evaluations cap. Finally, each experiment was repeated for $R=30$ independent runs, each time using a different training set. The training set for each experimental run was determined by sampling one-hundredth of a random shuffle of the 267\,292 headlines with six words. Thus, each training set was composed of $T=2\, 672$ sentences. 

\subsection{Training Phase Results}
\label{subsec:train}
\begin{figure}[t]
	\centering
	\includegraphics[scale=0.57]{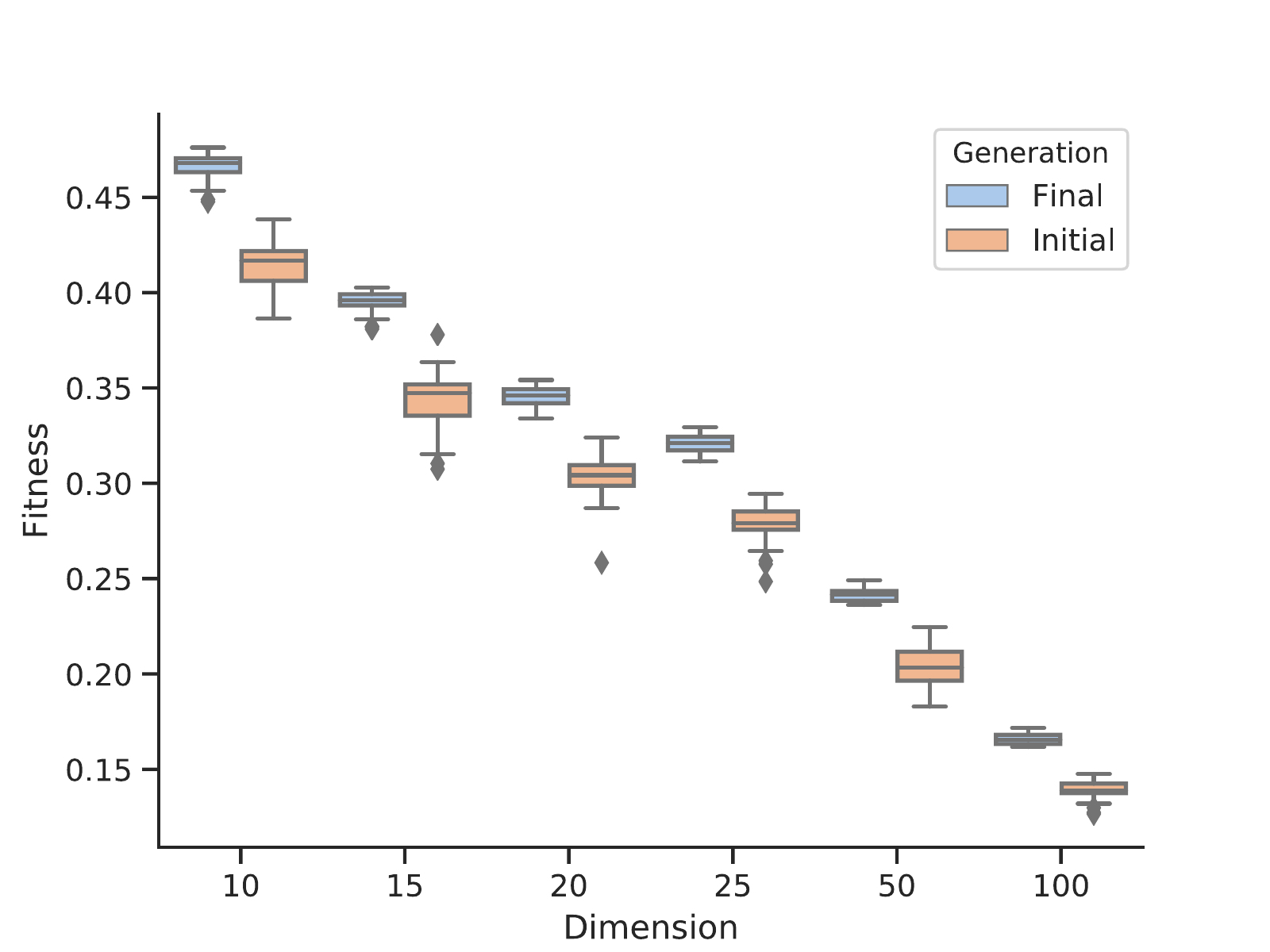}
	\caption{Best GP fitness at the first and last generation.}
	\label{fig:init-final}
\end{figure}

\begin{figure}[t]
	\centering
	\includegraphics[scale=0.57]{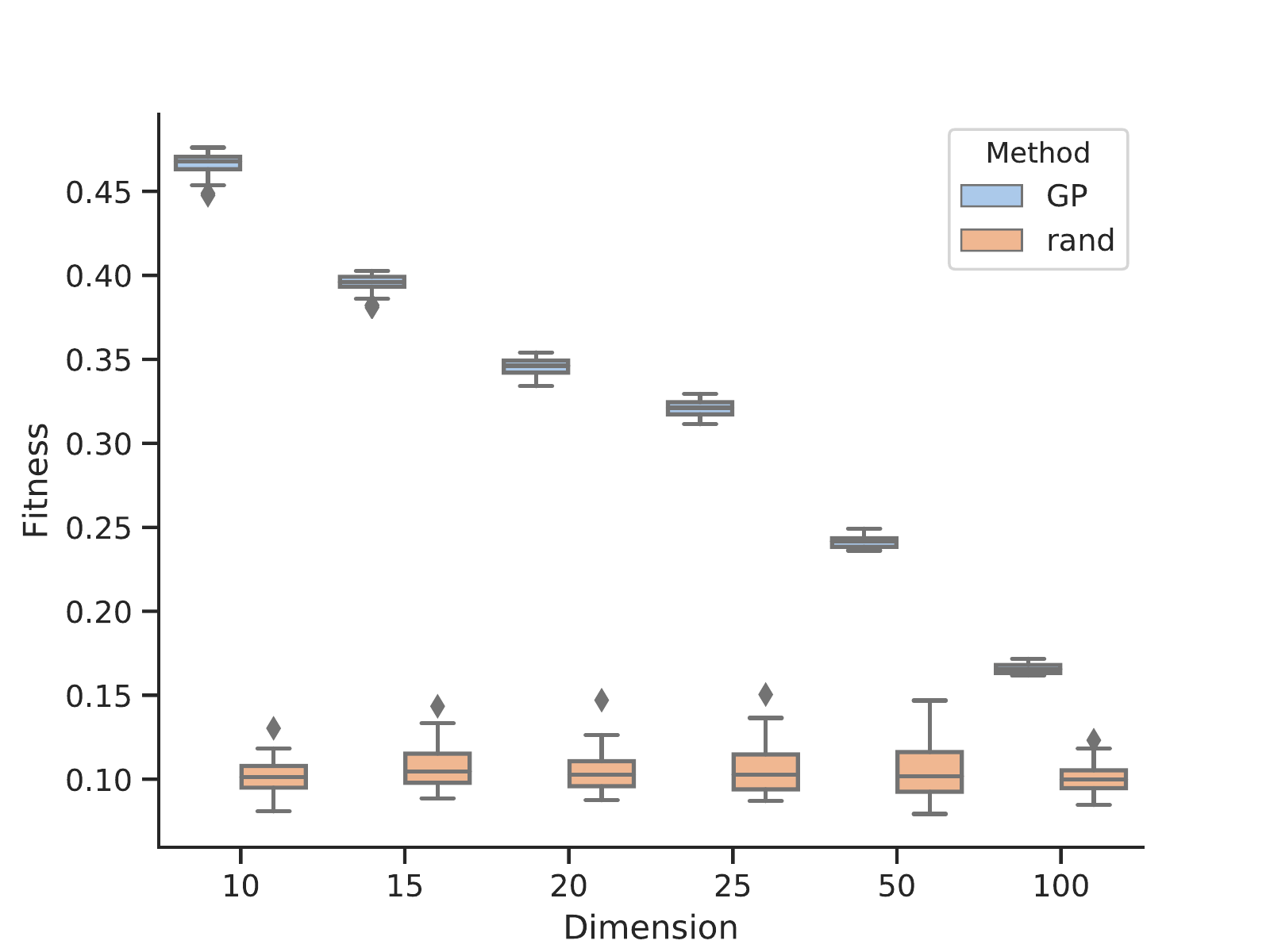}
	\caption{Best GP and random generation fitness values.}
	\label{fig:gp-rand}
\end{figure}

The first question that we addressed in our training experiments is whether the trees evolved by GP actually learned a model of the MNH dataset, under the embeddings produced by word2vec. To this end, for each embedding dimension, we compared the following distributions approximated by our $R=30$ experimental runs:
\begin{itemize}
	\item The fitness value of the best GP individual in the population before the first generation takes place (i.e., after randomly initializing the population) and at the end of the experimental run (Figure~\ref{fig:init-final}).
	\item The fitness value of the best GP individual at the end of the experimental run and the fitness value of the best random predictor (Figure~\ref{fig:gp-rand}).
\end{itemize}
In particular, the best random predictor was determined as follows. Given a training set of $T=2\, 672$ six-words headlines, for each embedding dimension $d \in \{10,15,20,25,50,100\}$ we generated $P=500$ sets of $T$ random vectors, and used them to predict the sixth word of the headlines. Stated differently, we employed these vectors as random surrogates for the outputs of the GP individuals over the training set. Subsequently, we scored each of the $P$ sets of vectors by applying the same fitness function used for GP, and the set achieving the highest fitness was selected as the best random predictor. We repeated this experiment for $R=30$ times, using the same training sets employed for the GP experimental runs. Remark that the best random predictors just described are different from the best individuals obtained by randomly initializing the GP population: indeed, the latter are GP trees that, although having a random structure, read the first five words of a sentence to predict the next one, while the former are random vectors completely independent from the content of the sentences.

The boxplots of Figures~\ref{fig:init-final} and~\ref{fig:gp-rand} clearly confirm that the trees evolved by GP are learning a model of the training sets considered in the experimental runs. As a matter of fact, for all considered dimensions, the fitness of the final best individual is always higher than that of both the best individual after initialization and the best random predictor. More specifically, the difference between the final best individual and the random predictor is much more significant, as Figure~\ref{fig:gp-rand} shows.

The second question that we investigated was whether a correlation exists between the dimension $d$ of the embedding space produced by word2vec and the fitness of the final best individual bred by GP. In particular, one can see from Figure~\ref{fig:gp-rand} that the lower the dimension of the embedding, the higher is the fitness attained by the best individual in the population. Although this finding is in stark contrast with the common NLP practice based on deep learning models, where the involved word2vec embeddings usually have hundreds of dimensions, this is a reasonable outcome for our GP algorithm. In fact, one can expect that a GP tree has an easier time in overfitting a training set of a couple of thousands of samples over a lower-dimensional space rather than on a higher-dimensional one since we enforce a maximum depth of the tree equal to the number of input words. Considering the relatively small size of our text corpus, it could be the case that the word2vec embeddings with lower dimensions produce vectors that are more closely packed, thus making overfitting easy for a compact GP tree. On the other hand, for the embeddings of higher dimension, the resulting vectors could be sparser, and thus harder to handle for small trees. 

\begin{figure}[t]
	\centering
	\includegraphics[scale=0.57]{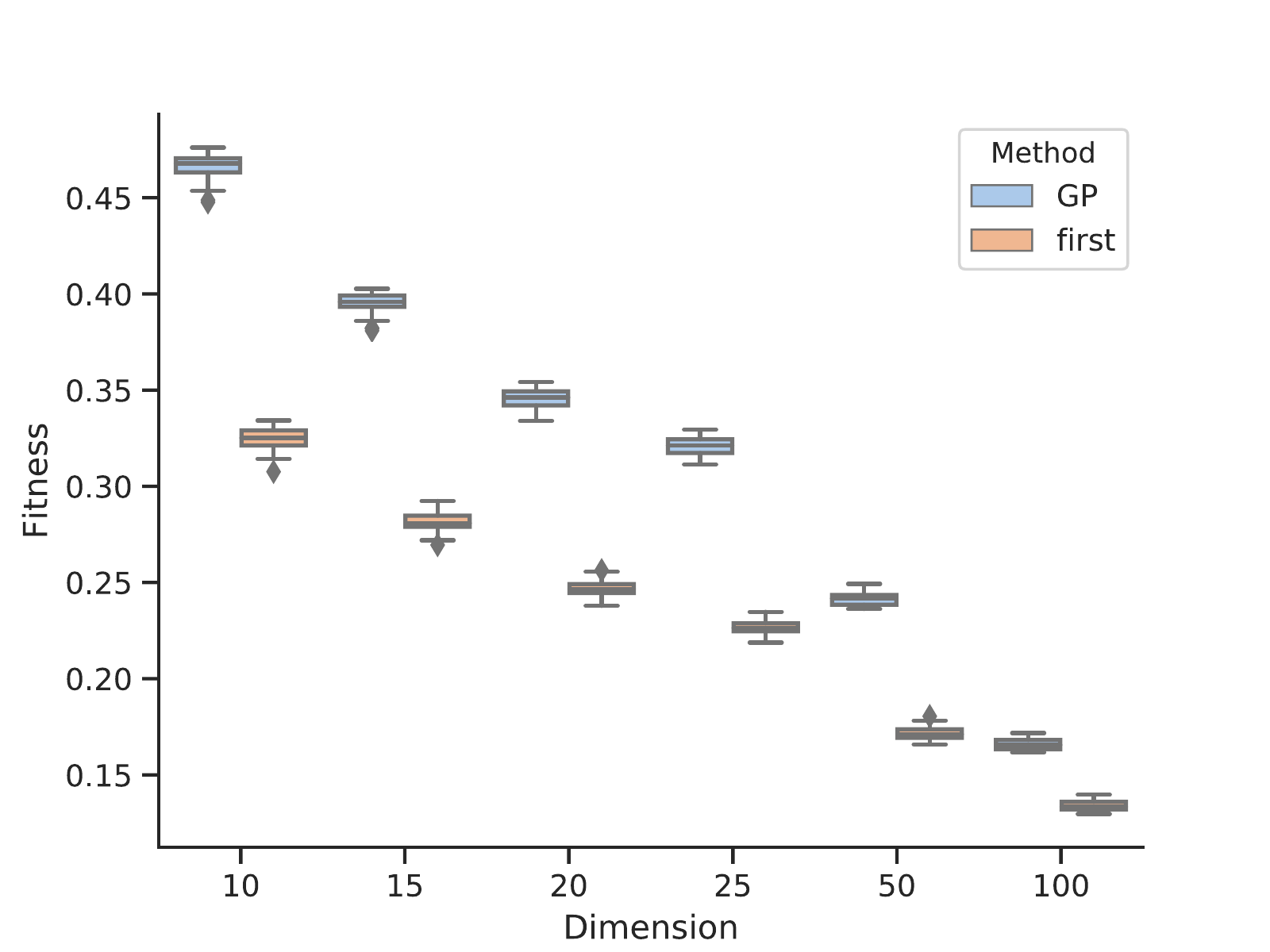}
	\caption{Best GP and "first-predictor" fitness values.}
	\label{fig:gp-first}
\end{figure}
\begin{figure}[t]
	\centering
	\includegraphics[scale=0.57]{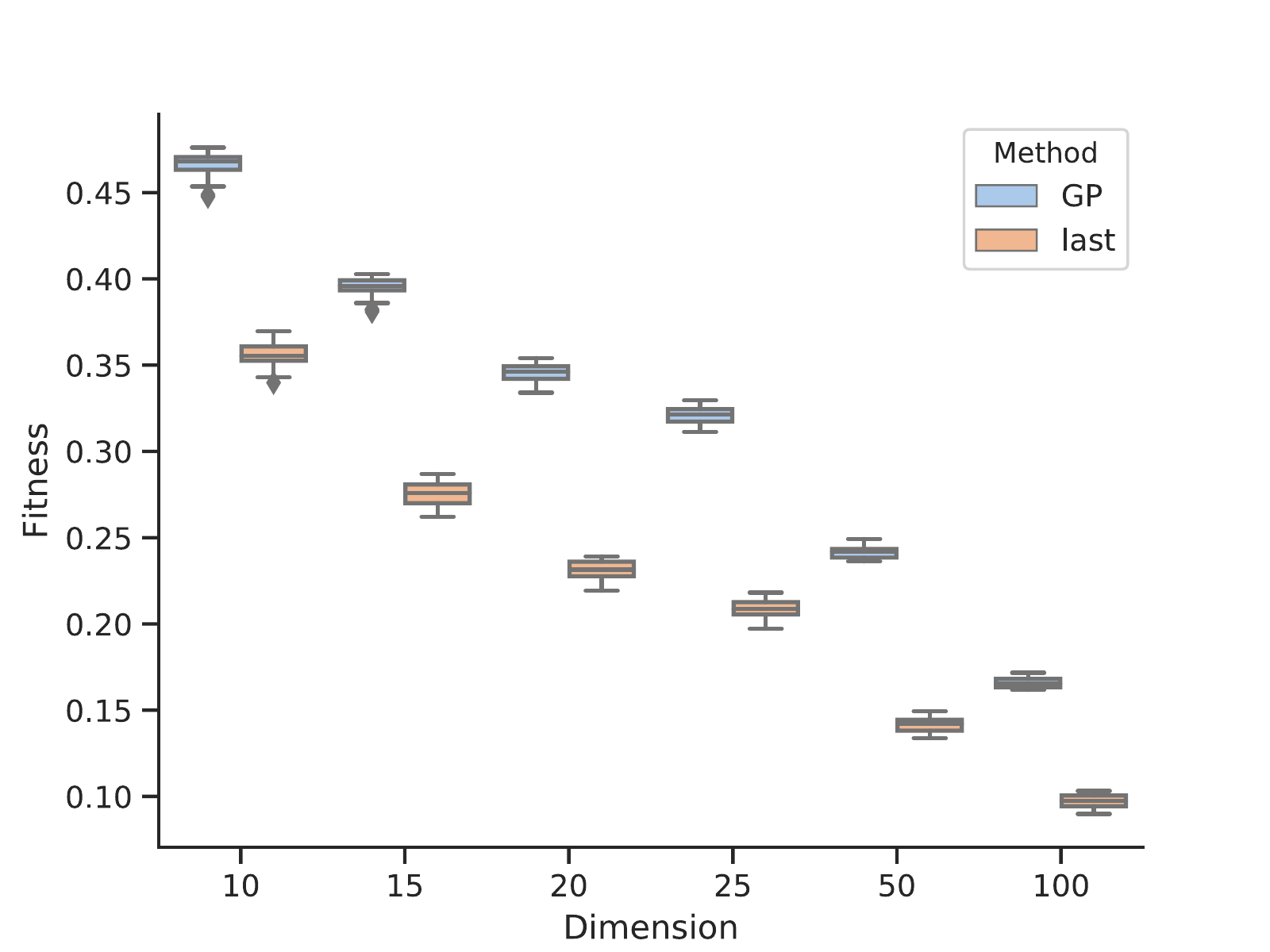}
	\caption{Best GP and "last-predictor" fitness values.}
	\label{fig:gp-last}
\end{figure}

At first glance, this overfitting hypothesis seems to indicate that the GP trees evolved over lower dimensional embeddings fare worse on the test set. For this reason, we performed a preliminary validation test on a small random sample of 50 headlines of six words. Surprisingly, we remarked that the best GP trees evolved over the higher dimensional embeddings (i.e., $d=50, 100$) resulted in a completely uninteresting behavior, since they learned to predict only one of the first five words seen in input to complete the sentence, almost always the first one. On the contrary, although the GP trees trained with the embeddings of dimension $10, 15, 20$, and $25$ failed to predict the original target words in the sample, nevertheless, they completed the sentences more creatively, producing plausible headlines in some cases. To further corroborate this observation, we compared the distribution of the best individuals evolved by GP with the trivial generators that always predict the first and the fifth word of a sentence, using the same training sets adopted for GP. For each considered dimension, Figures~\ref{fig:gp-first} and~\ref{fig:gp-last} report the boxplots of the fitness values attained by the best GP trees and the "first-word" and "last-word" predictors, respectively.
One can notice from Figure~\ref{fig:gp-first}, that the boxplots for the ``first-word'' predictor and the best GP individual obtained over dimension $d=100$ are quite close to each other, thereby providing further evidence that this is the behavior learned by those GP trees over the higher dimensional embeddings. On the contrary, there is a clear difference between the distribution of the best GP individuals and the first-word predictor for lower dimensions, showing that the fitness of the former is much higher than that scored by the latter. Hence, this seems to indicate that allowing to overfit the training data over a lower-dimensional embedding is beneficial to evolve GP trees that predict meaningful completions.

A final remark that can be drawn from the previous plots is that the distributions of the best fitness values obtained by GP at the final generation are not widely dispersed. Indeed, especially for high dimensions, the first and third quartiles are quite close to the median best fitness. The only remarkable difference is in the lower dimensional embeddings, where there is a wider difference between the minimum and the maximum best fitness. Nonetheless, this seems to indicate that the best GP trees evolved over a specific embedding learn the same prediction model. This hypothesis is also supported by our preliminary validation test, since for each dimension $d$ the best individuals obtained over the $R=30$ runs completed the 50 headlines approximately in the same way, almost always by predicting the same word for each sentence.

\subsection{Testing Phase Results}
\label{subsec:valid}
\begin{figure}[t]
	\centering
	\begin{tikzpicture}[scale=0.9]
	\tikzset{every tree node/.style={draw, thick, circle,inner sep=1pt, minimum size=0.7cm}}
	\tikzset{every leaf node/.append style={fill=black!10}}
	\tikzset{edge from parent/.append style={thick}}
	\Tree [.$+$
	[.$+$ [.$+$ $w_2$ [.+ $w_0$ $w_4$ ] ] 
	[.$+$ [.$+$ $w_1$ $w_4$ ] [.$(\cdot)^2$ $w_2$ ] ] ]
	[.$-$ [.$+$ $w_3$ [.$\sqrt{\cdot}$ $w_0$ ] ]
	[.$\times$ [.$-$ [.$\sqrt{\cdot}$ $w_2$ ]
	[.$\times$ $w_1$ $w_4$ ] ] [.$\sqrt{\cdot}$ $w_2$ ] ] ]
	]
	\end{tikzpicture}
	\caption{GP tree of the best individual evolved for dimension $d=10$ during one of the runs.}
	\label{fig:example-tree}
\end{figure}
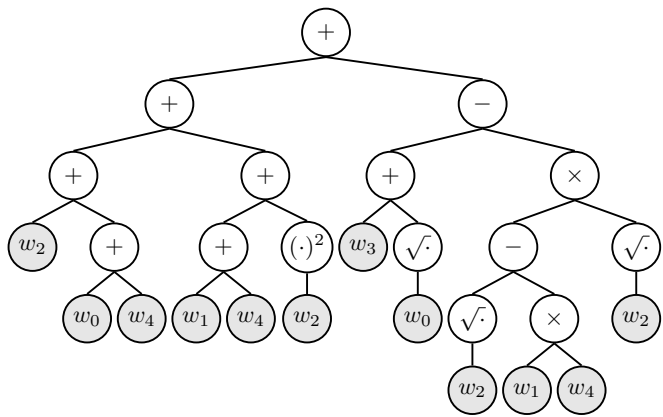

To better investigate the prediction models learned by the trees evolved with GP over the training set, we performed a more systematic testing phase designed as follows. For each of the considered six embedding dimensions, we selected the best GP tree with the highest fitness among the $30$ experimental runs. Table~\ref{tab:best-ind} reports the algebraic expressions of the best-selected individuals together with their sizes, while  Figure~\ref{fig:example-tree} depicts the tree of the best-selected individual for dimension $d=10$ as an example. Notice that in Table~\ref{tab:best-ind}, we displayed the non-simplified expression of the trees since the output of each internal node undergoes normalization, which would have burdened the notation if we included it in the simplified formulae.
\begin{table}[t]
	\centering
	\begin{tabular}{ccl}
		\hline\noalign{\smallskip}
		$d$ & Size & Expression  \\
		\noalign{\smallskip}\hline
		\noalign{\smallskip}\hline
		$10$  & $27$ & $(((w_2+(w_4+w_0))+((w_4+w_1)+w_2^2))+$ \\
		&      & $((w_3+w_0^2)-((\sqrt{w_2}-(w_1\cdot w_4))\cdot \sqrt{w_2}))))$ \\
		\noalign{\smallskip}
		$15$  & $38$ & $(((((w_4+w_0)+(w_3+w_2))+(w_1+(w_2\cdot w_2)))+$ \\
		&      & $(w_3+(w_4+w_4)))+ ((((w_0+w_0)+w_2)-$ \\
		&      & $(w_4\cdot (w_3-w_4)))+(w_1+(\sqrt{w_3}\cdot w_3))))$ \\
		\noalign{\smallskip}
		$20$  & $39$ & $(((((w_3+w_0)+w_0^2)+(w_1+w_1))+(w_1+w_4))+$ \\
		&      & $((((w_4+w_0)+(w_2+w_0))+(w_4+w_2))+$ \\
		&      & $(((w_3+w_0)+ w_2^2)+(w_4+(w_4\cdot w_4)))))$\\
		\noalign{\smallskip}
		$25$  & $48$ & $(((((w_4\cdot w_4)+(w_4+w_0))+((w_1+w_3)+$ \\
		&      & $(w_4+w_2)))+((w_3^2+w_1)+(w_0+(w_0+w_4))))+$ \\
		&      & $((((w_1+w_3)+(w_4+w_0))+(w_2+w_0))+$ \\
		&      & $(((w_2-w_1)\cdot (w_2-w_1))+w_2)))$ \\
		\noalign{\smallskip}
		$50$  & $36$ & $((((w_4+(w_3+w_0))+(w_1-w_3^2))+((w_4+$ \\
		&      & $(w_0+w_1))+w_2))+((w_2+(w_4+w_0))+$ \\
		&      & $((w_3+w_0)+((w_2-w_3)\cdot (w_1+w_3)))))$\\
		\noalign{\smallskip}
		$100$ & $27$ & $((((w_0+(w_3\cdot w_3))+w_1^2)+(w_2+w_1))+$ \\
		&      & $((((w_4+w_0)+w_3)+w_1^2)-(w_4\cdot (w_1-w_4))))$\\
		\noalign{\smallskip}
		\hline
		\smallskip
	\end{tabular}
	\caption{Algebraic expressions and sizes of the best GP individuals tested for each embedding dimension.}
	\label{tab:best-ind}
\end{table}
Next, we tested each selected individual over a random test sample of 10\,000 headlines of six words from the MNH dataset. Analogously to the preliminary validation step described in the previous section, we evaluated the GP tree over the embedding vectors corresponding to the first five words for each sentence in the test set, thus obtaining a sixth vector in output. Since this output vector does not correspond in general to a word of our corpus under the considered word2vec embedding, the predicted word is the one whose embedding vector has the highest cosine similarity with the output vector. Finally, for each sentence, we computed the cosine similarity between the predicted word and the original sixth word. Cosine similarity of $1$ thus means that the GP tree predicts exactly the original sixth word of a headline. However, remark that predicting the original word is not the main task of the models encoded by the GP trees since each headline can be completed in many ways that are both syntactically and semantically valid. In general, we observed that high values of cosine similarity usually correspond to meaningful completions of the headlines. 

\begin{figure}[t]
	\centering
	\includegraphics[scale=0.51]{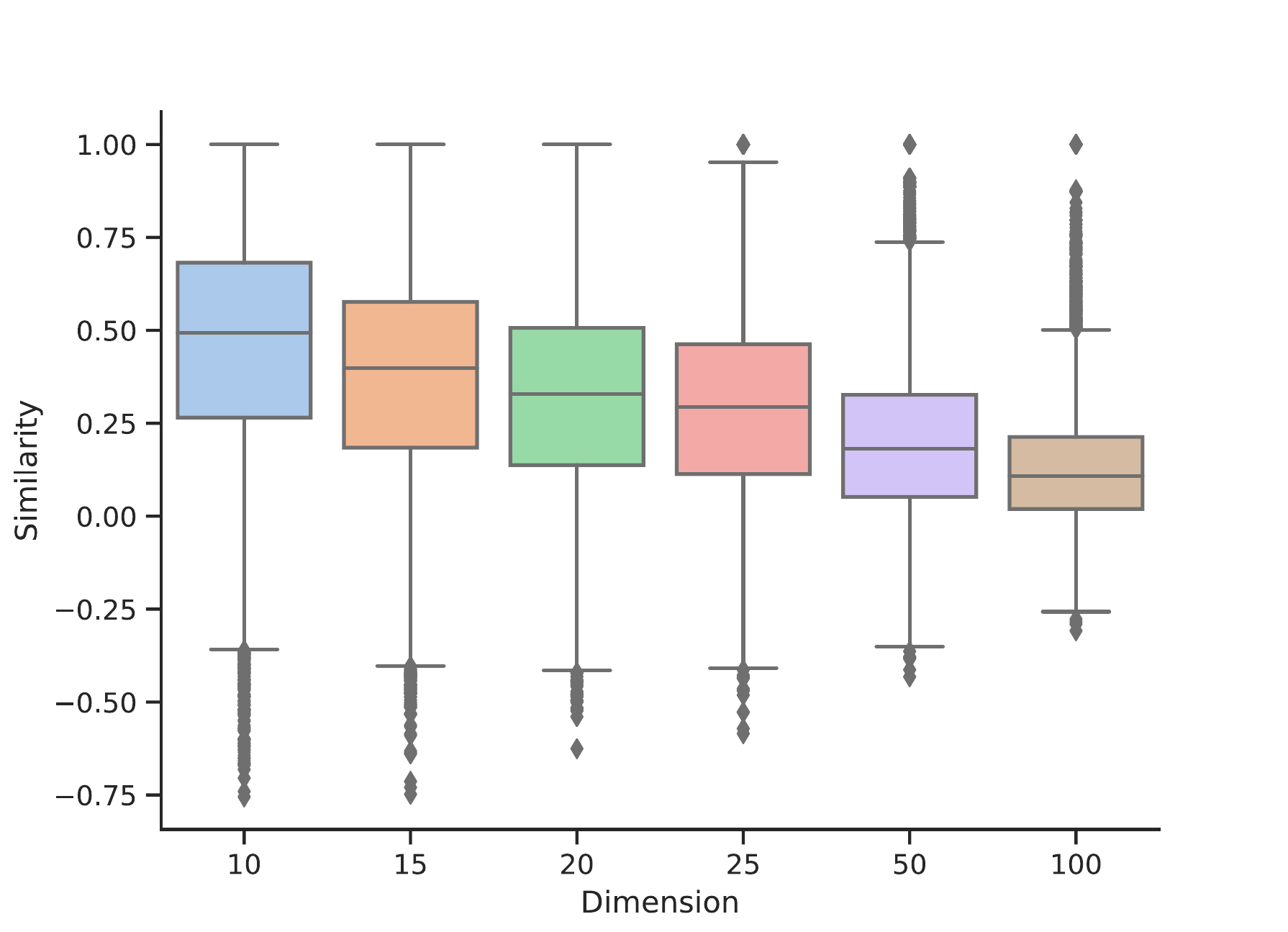}
	\caption{Distributions of similarity between predicted and original word over the test set.}
	\label{fig:sim-test}
\end{figure}

Figure~\ref{fig:sim-test} reports the boxplots of the distributions of similarity between predicted and original word over the test sample of 10\,000 headlines.
One can see from the plots that the initial findings suggested by the training phase results are confirmed. Indeed, the cosine similarity between predicted and original words obtained by the GP trees trained over higher-dimensional embeddings is lower than the similarity achieved by trees that were evolved over the lower-dimensional embeddings. In particular, it can be remarked that the outliers are arranged rather symmetrically: while for lower dimensions they are all focused below the bottom whisker, for dimension $d=50$ and $d=100$ they are almost all above the top one. On the other hand, the GP trees trained over dimension up to $20$ consistently reach higher values of cosine similarity.

As we remarked in the previous section, the GP individuals predicting a final word having a high cosine similarity with the original word often resulted in meaningful headlines, when considering a small test set of 50 sentences, and we observed the same behavior also over the larger test set of $10\, 000$ sentences. Table~\ref{tab:ex-sentences} reports a small selection of headlines completed by the best GP individual of dimension $10$ on the test set, together with the original word and the associated cosine similarity. In particular, we chose among predictions that reached a cosine similarity of at least $0.8$.
\begin{table}[t]
	\centering
	\begin{tabular}{ll}
		\hline\noalign{\smallskip}
		Predicted headline & Original  \\
		\noalign{\smallskip}\hline
		\noalign{\smallskip}\hline
		Regional education to fund youth \emph{preschool} & allowance  \\
		Aerial footage of flooded Townsville \emph{houses} & homes \\
		Greens renew call for tax \emph{changes} & review \\
		Napthine to launch new Portland \emph{rail} & marina \\
		4 charged over 10000 jewellery \emph{robberies} & heist \\
		Vanstone defends land rights act \emph{overhaul} & changes \\
		Community urged to seek infrastructure \emph{funds} & funding \\
		Govt. pressured on company tax \emph{bureaucracy} & rates  \\
		Petition urges probe into abattoir \emph{maintenance} & closure \\
		Rain does little for central \emph{towns} & Victoria \\
		\noalign{\smallskip}\hline
	\end{tabular}
	\caption{Examples of test headlines completed by the best GP individual of dimension $10$.}
	\label{tab:ex-sentences}
\end{table}
It can be seen from the table that in most cases the meaning of the original sentence is completely changed by the prediction. Indeed, only for three sentences the predicted word is a synonym of the original one (see \emph{houses}--\emph{homes}, \emph{robberies}--\emph{heist} and \emph{funds}--\emph{funding}). However, all resulting headlines are plausible, and one can observe that the predicted word is always related to a similar context with respect to the original word. For example, in the last example of Table~\ref{tab:ex-sentences} the predicted word refers to a generic place (\emph{towns}), while the original final word is the name of a region (\emph{Victoria}). Moreover, the third and the eight examples suggest that the model learned by the GP tree is also able to discriminate between different contexts when the same word occurs in the input. In fact, in the third sentence, the word \emph{tax} has been matched with \emph{changes} instead of \emph{review}, hence preserving the political overtone of the headline. Contrarily, in the eight sentence the GP tree paired \emph{tax} with \emph{bureaucracy} in place of \emph{rates}, thus remaining on a more administrative/economic context.


\section{Discussion}
\label{sec:disc}
The results presented in the previous two sections suggest that, to some extent, GP can learn a language model from a text dataset by leveraging the embedding produced by word2vec. We now summarize our main findings, framing them in a critical perspective to address the advantages and the shortcomings of the approach proposed in this paper.

\subsubsection*{Learning vs. Exact Prediction}
We have observed that GP does not usually predict the exact missing word in a sentence, i.e., an \emph{exact prediction}. However, we do not consider this a significant problem. In fact, there are multiple reasonable ways to complete sentences, and a dataset might itself contain multiple endings for the same sentence. What we are more interested in is the ability of GP to produce a \emph{meaningful} completion of a sentence. This ability is, in some sense, more interesting. It shows that GP has learned to ``navigate'' the space given by the word2vec embedding and produce results that align with the semantics of the sentences.

\subsubsection*{Dimensionality and Fitness}
One interesting observation from the experiments is that the dimensionality of the embedding has a large influence on the behavior of GP. First of all, the increase in the fitness values can be partially dictated purely by geometrical effects; in fact, the average fitness for the trivial predictors (that predict only the first or last word of a sentence) also increases. However, the increase in fitness obtained by GP is higher, showing that lower-dimensional spaces might be easier to manipulate by this kind of GP. Furthermore, the size of the embeddings used in this work is quite small compared to the ones usually employed, which can have up to hundreds of dimensions. The motivation for GP preferring lower-dimensional spaces is an interesting topic for future investigations.

\section{Conclusions and Future Works}
\label{sec:conc}

In this paper, we introduced a new approach to perform an NLP task, namely next word prediction, using GP combined with the word embeddings produced by word2vec. This combination required a way to manipulate vectors via GP (both as inputs and as outputs), to subsequently be able to ``go back'' from the embedding and generate words via GP. We have tested our approach on The Million News Headlines dataset. To check that the GP trees were not learning only trivial functions of the input, we have compared their fitness with random and trivial predictors, showing that GP is actually learning a model of the language represented by the dataset. By looking at how unseen sentences were completed, we found that, for dimension $10$, a median cosine similarity around $0.5$ was obtained, and, among the word generated, reasonable completions were usually obtained. While the results of this work do not top the current state of the art, our approach shows that GP is a promising technique for this kind of task, opening many research directions and improvements that, hopefully, will bring GP among the top techniques in NLP. In fact, we are only moving \emph{toward} a possible application of GP to NLP tasks. Here, we mention some possible directions for future research; the list is by no means exhaustive, but it should provide an ample spectrum of possible improvements.

We have only employed operators that are pointwise adaptations of the classical GP operators. An immediate improvement would be to add operators that are \emph{specific} for arrays of unitary length, such as \emph{rotations} and other linear transformations.

The GP trees employed are deterministic; if we have two sentences that agree everywhere except for the word to be predicted, it is impossible to generate them in a stochastic way. Therefore, an immediate extension is to make the generation \emph{probabilistic}. There are multiple ways to accomplish this; among them, one possibility is to use an entire set of trees for constructing the word and picking a single tree randomly each time a word needs to be generated.

We also tested our approach only on a single dataset of headlines, and we also restricted the training to a subset of it. It is essential to extend the approach to a larger number of datasets; this will require multiple improvements in terms of scalability of the GP algorithm, and possibly in terms of the management of low-frequency words, which could also be treated as ``unknown'' tokens.

While next word prediction has been the focus of this work, our approach can be extended to generate longer texts by using a ``sliding window'' methods: given the words $w_1,\ldots,w_k$, the word $w_{k+1}$ is predicted as usual; then, the words $w_2, \ldots, w_{k+1}$ are used to generate $w_{k+2}$, and so on. Clearly, such an approach does not keep any internal state in the GP tree, and it leverages only on the last $k$ words. Therefore, it would be interesting to add a ``memory'' feature to GP, similarly to LSTM networks~\cite{sundermeyer2012lstm,ebner2017distributed}.

Finally, to evaluate the quality of the generated words or sentences, a competitive (or ``adversarial'') co-evolution could be employed, where there is a population of trees that act as the generators of sentences, while another population contains trees that act as discriminators between real and generated sentences.

\section*{Acknowledgments}
This work was partially supported by national funds through FCT (Funda\c{c}\~{a}o para a Ci\^{e}ncia e a Tecnologia) by the project GADgET (DSAIPA/DS/0022/2018) and AICE (DSAIPA/DS/0113/2019). Mauro Castelli acknowledges financial support from the Slovenian Research Agency (Javna Agencija za Raziskovalno Dejavnost RS) under the research core funding No.\ P5-0410. Luca Mariot acknowledges financial support from COST Action CA15140 -- Improving Applicability of Nature-Inspired Optimisation by Joining Theory and Practice (ImAppNIO).

\bibliographystyle{abbrv}
\bibliography{references}

\begin{thebibliography}{10}

\bibitem{al2008application}
H.~Al-Mubaid and P.~Chen.
\newblock Application of word prediction and disambiguation to improve text
  entry for people with physical disabilities (assistive technology).
\newblock {\em International Journal of Social and Humanistic Computing},
  1(1):10--27, 2008.

\bibitem{alshemali2019improving}
B.~Alshemali and J.~Kalita.
\newblock Improving the reliability of deep neural networks in nlp: A review.
\newblock {\em Knowledge-Based Systems}, page 105210, 2019.

\bibitem{araujo2004genetic}
L.~Araujo.
\newblock Genetic programming for natural language parsing.
\newblock In {\em European Conference on Genetic Programming}, pages 230--239.
  Springer, 2004.

\bibitem{1010071184429744}
L.~Araujo.
\newblock Multiobjective genetic programming for natural language parsing and
  tagging.
\newblock In T.~P. Runarsson, H.-G. Beyer, E.~Burke, J.~J. Merelo-Guerv{\'o}s,
  L.~D. Whitley, and X.~Yao, editors, {\em Parallel Problem Solving from Nature
  - PPSN IX}, pages 433--442. Springer, 2006.

\bibitem{araujo2019genetic}
L.~Araujo.
\newblock Genetic programming for natural language processing.
\newblock {\em Genetic Programming and Evolvable Machines}, pages 1--22, 2019.

\bibitem{banzhaf2019genetic}
W.~Banzhaf, R.~S. Olson, W.~Tozier, and R.~Riolo.
\newblock {\em Genetic Programming Theory and Practice XVI}.
\newblock Springer, 2019.

\bibitem{bartoli2014automatic}
A.~Bartoli, G.~Davanzo, A.~De~Lorenzo, E.~Medvet, and E.~Sorio.
\newblock Automatic synthesis of regular expressions from examples.
\newblock {\em Computer}, 47(12):72--80, 2014.

\bibitem{bartoli2017active}
A.~Bartoli, A.~De~Lorenzo, E.~Medvet, and F.~Tarlao.
\newblock Active learning of regular expressions for entity extraction.
\newblock {\em IEEE transactions on cybernetics}, 48(3):1067--1080, 2017.

\bibitem{bartoli2015evolutionary}
A.~Bartoli, A.~De~Lorenzo, E.~Medvet, F.~Tarlao, and M.~Virgolin.
\newblock Evolutionary learning of syntax patterns for genic interaction
  extraction.
\newblock In {\em Proceedings of the 2015 Annual Conference on Genetic and
  Evolutionary Computation}, pages 1183--1190, 2015.

\bibitem{dale2000handbook}
R.~Dale, H.~Moisl, and H.~Somers.
\newblock {\em Handbook of natural language processing}.
\newblock CRC Press, 2000.

\bibitem{de2010genetic}
M.~G. De~Carvalho, A.~H. Laender, M.~A. Gon{\c{c}}alves, and A.~S. da~Silva.
\newblock A genetic programming approach to record deduplication.
\newblock {\em IEEE Transactions on Knowledge and Data Engineering},
  24(3):399--412, 2010.

\bibitem{ebner2017distributed}
M.~Ebner.
\newblock Distributed storage and recall of sentences.
\newblock {\em Bio-Algorithms and Med-Systems}, 13(2):89--101, 2017.

\bibitem{garay1994using}
N.~Garay and J.~Abascal.
\newblock Using statistical and syntactic information in word prediction for
  input speed enhancement.
\newblock {\em Information Systems Design and Hypermedia}, pages 223--230,
  1994.

\bibitem{garay2006text}
N.~Garay-Vitoria and J.~Abascal.
\newblock Text prediction systems: a survey.
\newblock {\em Universal Access in the Information Society}, 4(3):188--203,
  2006.

\bibitem{garay1997intelligent}
N.~Garay-Vitoria and J.~Gonzalez-Abascal.
\newblock Intelligent word-prediction to enhance text input rate (a syntactic
  analysis-based word-prediction aid for people with severe motor and speech
  disability).
\newblock In {\em Proceedings of the 2nd international conference on
  Intelligent user interfaces}, pages 241--244, 1997.

\bibitem{isele2013active}
R.~Isele and C.~Bizer.
\newblock Active learning of expressive linkage rules using genetic
  programming.
\newblock {\em Journal of web semantics}, 23:2--15, 2013.

\bibitem{kim2004user}
K.-M. Kim, S.-S. Lim, and S.-B. Cho.
\newblock User adaptive answers generation for conversational agent using
  genetic programming.
\newblock In {\em International Conference on Intelligent Data Engineering and
  Automated Learning}, pages 813--819. Springer, 2004.

\bibitem{rohit18}
R.~Kulkarni.
\newblock {A Million News Headlines}, 2018.
\newblock Accessed on February 1st, 2020.

\bibitem{manurung2008implementation}
R.~Manurung, G.~Ritchie, and H.~S. Thompson.
\newblock An implementation of a flexible author-reviewer model of generation
  using genetic algorithms.
\newblock In {\em Proceedings of the 22nd Pacific Asia Conference on Language,
  Information and Computation}, pages 272--281, 2008.

\bibitem{mikolov2013}
T.~Mikolov, K.~Chen, G.~Corrado, and J.~Dean.
\newblock Efficient estimation of word representations in vector space.
\newblock In {\em {ICLR} (Workshop Poster)}, 2013.

\bibitem{poli08}
R.~Poli, W.~B. Langdon, and N.~F. McPhee.
\newblock {\em A Field Guide to Genetic Programming}.
\newblock lulu.com, 2008.

\bibitem{rose1999genetic}
C.~P. Ros{\'e}.
\newblock A genetic programming approach for robust language interpretation.
\newblock {\em Advances in genetic programming}, 3:67--88, 1999.

\bibitem{sundermeyer2012lstm}
M.~Sundermeyer, R.~Schl{\"u}ter, and H.~Ney.
\newblock Lstm neural networks for language modeling.
\newblock In {\em Thirteenth annual conference of the international speech
  communication association}, 2012.

\bibitem{swiffin1987use}
A.~L. Swiffin, J.~L. Arnott, and A.~F. Newell.
\newblock The use of syntax in a predictive communication aid for the
  physically handicapped.
\newblock In {\em 10th Annual Conference on Rehabilitation Technology}, pages
  124--126. RESNA-Association for the Advancement of Rehabilitation Technology,
  1987.

\bibitem{venkatagiri1993efficiency}
H.~Venkatagiri.
\newblock Efficiency of lexical prediction as a communication acceleration
  technique.
\newblock {\em Augmentative and Alternative Communication}, 9(3):161--167,
  1993.

\bibitem{young2018recent}
T.~Young, D.~Hazarika, S.~Poria, and E.~Cambria.
\newblock Recent trends in deep learning based natural language processing.
\newblock {\em {IEEE} Computational intelligence magazine}, 13(3):55--75, 2018.

\end{thebibliography}

\end{document}